%% file: main.tex
\documentclass{article}

\usepackage[final]{colm2026_conference}

\usepackage[utf8]{inputenc}
\usepackage[T1]{fontenc}
\usepackage{hyperref}
\usepackage{url}
\usepackage{booktabs}
\usepackage{amsfonts}
\usepackage{amsmath}
\usepackage{amssymb}
\usepackage{amsthm}
\usepackage{microtype}
\usepackage{xcolor}
\usepackage{graphicx}
\usepackage{multirow}
\usepackage{colortbl}
\usepackage{enumitem}
\usepackage{listings}
\usepackage{lineno}
\definecolor{noiseband}{HTML}{EAF3F9}%
\definecolor{biasband}{HTML}{FDECEA}%
\definecolor{judgeband}{HTML}{F0EAF6}%

\definecolor{darkblue}{rgb}{0, 0, 0.5}
\hypersetup{colorlinks=true, citecolor=darkblue, linkcolor=darkblue, urlcolor=darkblue}

\title{The Blind Curator: How a Biased Judge Silently Disables\\
Skill Retirement in Self-Evolving Agents}

\author{Xing Zhang$^{1}$, Yanwei Cui$^{1}$, Guanghui Wang$^{1}$, Ziyuan Li$^{2}$, Wei Qiu$^{2}$, \\
\bf Bing Zhu$^{2}$, Peiyang He$^{1}$\thanks{Corresponding author: \texttt{peiyan@amazon.com}} \\[3pt]
$^{1}$AWS Generative AI Innovation Center \\
$^{2}$HSBC Holdings Plc., HSBC Technology Center, China
}

\begin{document}

\ifcolmsubmission
\linenumbers
\fi

\maketitle

\begin{abstract}
A self-evolving agent retires its bad skills by watching them fail,
so what happens when the judge cannot see the failures? Skill
retirement is the structural constraint that keeps a growing library
from drifting below the no-skill baseline, but its guarantee assumes
an unbiased reward, which is false for the LLM judges that
reference-free tasks require. We show that a biased judge does
not merely add noise; it \emph{silently switches off the curator}. We
make this precise with a corrupted-reward analysis, then a behavioral
study on a reference-free report-writing testbed with a
code-generation cross-check, injecting corruption on top of a
deterministic reward to isolate the causal channel. Symmetric noise
leaves retirement intact, but \emph{false-pass} bias (failures
slipping through as passes) disables contribution-based retirement past
a sharp threshold (here a false-pass rate of $0.45$) that no amount of
data can cross. Separating genuine retirement from cap-eviction churn shows this
\emph{mechanism} failure is universal, holding across domains and
failure rates and sparing only near-zero-false-pass, verifier-like
graders. The downstream \emph{outcome}, though, is regime-dependent:
eval quality degrades only where the same corruption also starves skill
synthesis, and otherwise holds steady, so the disabled curator is
\emph{silent}, surfacing in no aggregate metric. The contribution is a
behavioral safety result, not a performance one. A cheap
defect-injection audit then tells an operator, before deployment, which
side of the threshold their judge occupies.
\end{abstract}

\section{Introduction}
\label{sec:intro}

\begin{figure}[tbp]
\centering
\includegraphics[width=\textwidth]{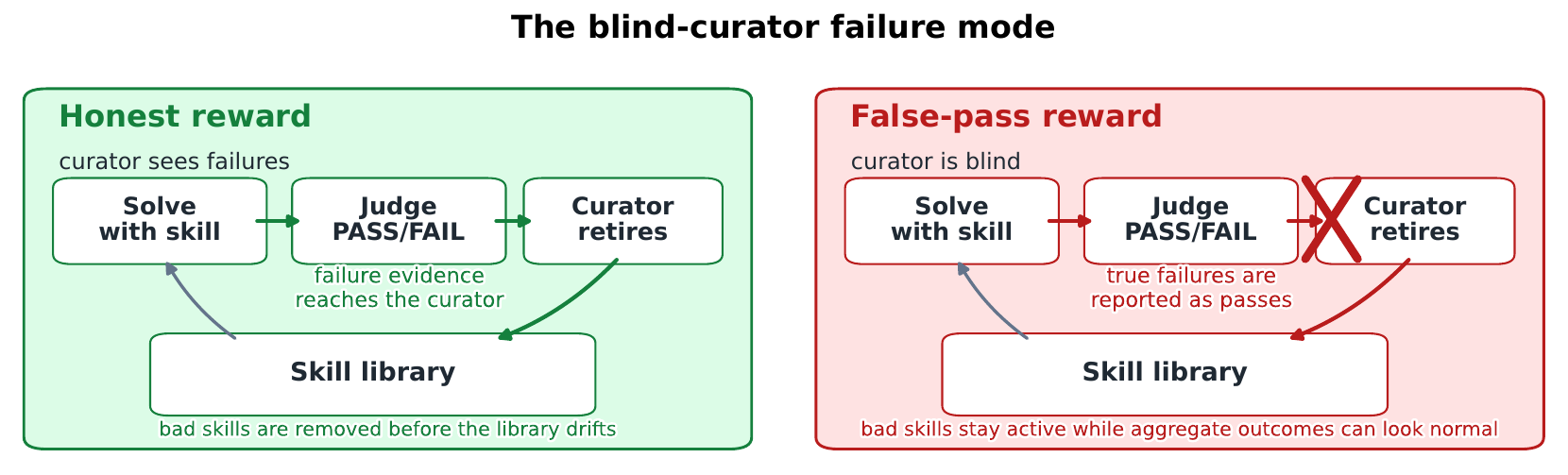}
\caption{The \emph{blind curator} failure mode. The same
failure-driven loop (solve, judge, retire) under an honest reward
(left) and a \emph{false-pass} judge (right): false passes break the
Judge$\to$Curator evidence, so retirement quietly stops while aggregate
outcomes can look normal. The gap opens at a sharp threshold.}
\label{fig:teaser}
\end{figure}

A self-evolving agent that accumulates skills without governance
degrades, as stale and redundant entries crowd retrieval, a failure
mode recently named \emph{library drift}~\citep{librarydrift2026}.
Agents already learn by synthesizing new skills from their own
failures~\citep{wang2023voyager, zhao2024expel}; the missing piece is
\emph{governance}, \emph{retiring} skills that stop helping under a
bounded cap. Ratchet~\citep{ratchet2026} makes this precise: retirement
keeps a growing library from drifting more than a fixed margin below
the no-skill baseline. But that guarantee rests on one quiet
assumption, that the signal telling the agent which skills failed is
\emph{honest} (the per-skill contribution estimator is unbiased).
Coding and question answering satisfy it with unit tests and
exact-match graders; the
tasks agents increasingly face (research synthesis, long-form
reporting, analysis) do not: with no golden answer the only scalable
grader is an LLM judge~\citep{zheng2023llmjudge}, and its error is not
white noise. Judges are systematically biased rather than merely
inconsistent~\citep{wang2024large, stureborg2024large}, tending to wave
through certain failure classes (a confident misquote; a flipped
conclusion that stays fluent), so their error is \emph{asymmetric}:
failures get reported as passes. The consequence is a \emph{blind
curator} (Fig.~\ref{fig:teaser}): the component that should retire bad
skills stops seeing the evidence it retires on. Library drift is the
disease governance was built to cure; curator blindness is what befalls
the cure when the reward is fallible.

We treat that asymmetry as the object of study. Our thesis turns on
\emph{two separable knobs} of the reward channel: a symmetric noise
rate $\rho$ (true label flipped either way), and a false-pass rate
$\rho_{F\to P}$, the fraction of true failures reported as passes
(we write $q$ for this same rate when sweeping it in the experiments).
The agent responds to them in opposite ways. It tolerates noise
gracefully, but hits a \emph{cliff} in $\rho_{F\to P}$: beyond
$\rho_{F\to P}=(1-\tau)/2$, under this channel, no amount of data can
rescue retirement (Sec.~\ref{sec:theory}). Our contributions are:

\begin{itemize}[leftmargin=1.2em,itemsep=2pt,topsep=2pt]
\item \textbf{A universal mechanism failure with regime-dependent
fallout.} Across four subsets and two domains
(Secs.~\ref{sec:sweep}--\ref{sec:crossdomain}), and by separating
genuine contribution-retirement from cap-eviction churn, we show
false-pass bias disables the \emph{curator} past the cliff in every
regime (verifier-like near-zero graders are spared; below the cliff
survival depends on skill margins), while the downstream \emph{outcome}
harm is regime-dependent: it appears only when the same corruption also
starves synthesis, and is otherwise silent. Symmetric noise is
survivable; a strict judge is safe but starvation-prone.
\item \textbf{A before-deployment go/no-go test
(Fig.~\ref{fig:playbook}).} A constructed-ground-truth testbed and a
defect-injection \emph{audit} that pinpoints any judge's error rates,
so an operator can locate their judge relative to the threshold before
trusting self-evolution (Sec.~\ref{sec:gate}).
\item \textbf{Why the asymmetry exists.} A bias-aware non-divergence
bound (Prop.~1$'$): noise only attenuates the retirement signal, while
false-pass bias displaces it past an unrecoverable threshold.
\end{itemize}

This is a behavior-centric study: we ask not whether task success
rises but \emph{how} the skill library evolves, and when retirement
under a cap reliably steers it versus silently failing. The audit is a
behavioral test suite for the reward; the blind curator is a
mechanistic account of a process-level failure that aggregate metrics
never surface.

\section{Related work}
\label{sec:related}

\textbf{Skill libraries and verbal self-improvement.}
A line of work lets a frozen LLM agent improve by accumulating
\emph{textual} artifacts from its own experience rather than updating
weights: reusable skills~\citep{wang2023voyager}, distilled
experience~\citep{zhao2024expel}, instruction
manuals~\citep{chen2024automanual}, episodic
reflections~\citep{shinn2023reflexion}, workflow
memory~\citep{wang2024awm}, and natural-language feedback as a
gradient surrogate~\citep{yuksekgonul2024textgrad, madaan2023selfrefine}.
A second wave strengthens skill creation itself, via inductive
distillation from traces~\citep{alibaba2026trace2skill}, autonomous
create-and-evolve loops~\citep{huang2025cascade, yang2026autoskill},
and RL-coupled skill growth~\citep{xia2026skillrl}, with benchmarks
probing transfer~\citep{li2026skillsbench} and a lifecycle
view~\citep{wu2025evolver} also underlying OS-like agent
memory~\citep{packer2023memgpt}. A recurring weakness is that the library only
grows~\citep{librarydrift2026}, and almost all of this work assumes
the outcome signal driving accumulation is \emph{correct}; we ask what
happens when it is not.

\textbf{Skill lifecycle governance.} Closest to us is a line that
treats the skill set as something to be \emph{governed}, not just
grown: lifecycle governance from collection to
evolution~\citep{liu2026skillsvote}, dynamic lifecycle management under
agentic RL~\citep{shen2026dynamic}, and trajectory-driven
self-adaptation~\citep{yu2026skilladaptor, cui2026feedback}. All share the premise that
bad skills must be pruned, but all presume the pruning signal is
trustworthy. Our contribution is prior to the governance policy: we
characterize when the \emph{signal} any such policy consumes is
reliable enough for pruning to help rather than hurt.

\textbf{The Ratchet mechanism we build on.} We start from
Ratchet~\citep{librarydrift2026, ratchet2026}, a minimal recipe on a
frozen LLM: a Critic labels each failed task, a Synthesizer turns
recurring failure patterns into skills, a Router retrieves at most one
skill per task,
and, crucially, a Curator \emph{retires} any skill whose empirical
contribution falls below a threshold after enough trials, under a hard
cap on active skills. Retirement-plus-cap yields a
\emph{non-divergence} guarantee: on a fixed task distribution,
expected performance cannot fall more than a fixed margin below the
no-skill baseline. We adopt Ratchet because it is, to our knowledge,
the only self-evolving-skill scheme with such a guarantee, which makes
it the natural object on which to ask what an imperfect reward does to
that guarantee: its proof is precisely what an unreliable judge can
break. It was evaluated only with clean verifiers: unit tests on
MBPP+~\citep{austin2021mbpp, liu2023evalplus} and the SWE-bench Docker
harness~\citep{jimenez2023swebench}. We move it to the verifier-free
regime its guarantee was never tested in.

\textbf{LLM-as-judge reliability and noisy supervision.} The reward in
our regime is an LLM judge~\citep{zheng2023llmjudge}, the standard way
to turn AI feedback into a learning signal~\citep{bai2022constitutional}.
A large body of work shows such judges are systematically biased rather
than merely noisy: position, verbosity, and self-enhancement
effects~\citep{zheng2023llmjudge}, order-dependence that can flip a
verdict~\citep{wang2024large}, run-to-run
inconsistency~\citep{stureborg2024large}, and the broader caveats
collected in recent surveys~\citep{li2024llms}. Where a cheap
executable verifier can be learned~\citep{pezeshkpour2026autopyverifier}
our concern does not arise; our focus is precisely the regime where
none exists. We add not another reliability study but the
\emph{propagation} of a judge's error \emph{asymmetry} through a
specific learning mechanism. The
closest classical framing is class-conditional label noise, but here
the ``labels'' gate a \emph{lifecycle decision} (retirement) inside a
closed loop, so asymmetric noise compounds: unretired bad skills keep
being routed to, generating more corrupted evidence. The analogue of
catastrophic forgetting~\citep{kirkpatrick2017ewc} is not weight
overwriting but the silent retention of harmful skills a blind curator
can no longer prune.

\section{Setup: failure-driven evolution on report composition}
\label{sec:setup}

\textbf{Why this testbed.} The phenomenon only exists where the reward
is a fallible judge, i.e., on tasks with no golden answer, which rules
out the verifier-backed benchmarks (MBPP, SWE-bench) Ratchet was built
on. We need a task that is (i) genuinely reference-free, (ii) yet has an
\emph{objective} sub-signal we can treat as ground truth and corrupt,
and (iii) shows the visible/invisible failure split that makes a judge
both necessary and fallible. Long-form, citation-grounded report
writing fits: there is no gold report, but citation discipline is
checkable while source faithfulness is not. We draw tasks from an
internal production deep-research engine;
Sec.~\ref{sec:crossdomain} tests how far they carry to a
verifier-backed domain.

\textbf{Task and reward.} A trial composes one report section from a
frozen \emph{evidence slice} (its cards, metric values, and allowed
cross-references; Appendix~\ref{app:slice}), citing only slice cards,
annotating every numeral, and meeting a structural contract. The 155
slices come from five complete productions of that engine, a
Chinese-language corpus; the examples in this paper are English glosses.
Freezing the slices makes trials cheap (one LLM call) and
i.i.d.-replayable, which the retirement statistic needs. The reward is a deterministic grader of
five \emph{quality-control (QC)} checks (orphan citation, unregistered
metric, unsourced number, broken cross-reference, missing TL;DR); PASS $=$
zero violations. Because it is reference-free yet objective, we treat
it as ground truth $y$ and impose corrupted channels $\tilde y$ on top;
a defect is ``QC-visible'' or ``QC-invisible'' by whether these checks
catch it.

\textbf{Evolution loop.} We run the governance stack unmodified (the
solve--judge--retire loop sketched in Fig.~\ref{fig:teaser}). Skills
here are composition disciplines (citation-coverage habits, hedging
templates) injected as prompt guidance.

\textbf{Failures are the single input the whole loop runs on.} This is
the structural reason false-pass bias is so damaging, and it holds for
any failure-driven scheme, not just this one. Observed failures are the
\emph{only} signal feeding both halves of the loop: the Synthesizer
clusters them into new skills, and the Curator retires a skill from its
observed fail/pass tally. A false-pass judge ($\tilde y$ reports a true
failure as a pass) therefore does damage at the source: it (i) shrinks
the observed-failure pool, which (ii) starves \emph{synthesis} (fewer
failures to cluster) and (iii) inflates every skill's observed pass
rate, so the \emph{Curator} sees nothing crossing $-\tau$. The
opposite corruption (true pass reported as a fail) only injects
\emph{phantom} failures, extra fuel the loop tolerates as noise. The
asymmetry we study is thus built into the loop's reliance on failures,
which is exactly the signal a production deployment logs and acts on.

\textbf{Hard subset.} Because a competent composer passes most
sections at temperature $0.7$, we probe all 155 tasks $\times 3$ and
keep the 71 with at least one failure (split 43 train / 28 eval,
stratified by report); details in Appendix~\ref{app:setup}.

\section{Why bias is different from noise: theory in brief}
\label{sec:theory}

Ratchet's non-divergence guarantee assumes the per-skill contribution
estimator is unbiased: innocuous with a deterministic grader, false
with an LLM judge whose errors are \emph{structured}. Model the judge
as a binary channel on the true outcome $y\in\{0,1\}$ ($1{=}$pass):
$\Pr[\tilde y{=}1\mid y{=}0]=\rho_{F\to P}$ (a hidden failure) and
$\Pr[\tilde y{=}0\mid y{=}1]=\rho_{P\to F}$ (a phantom failure).
The Curator retires a skill once its observed pass rate drops to
$\pi_\tau:=(1-\tau)/2$. Writing $\bar p(s)$ for skill $s$'s
\emph{true} pass rate on the tasks routed to it, under the channel its
observed pass rate concentrates on $\kappa\,\bar p(s)+\rho_{F\to P}$,
with $\kappa:=1-\rho_{F\to P}-\rho_{P\to F}$, and the two corruption
types act on it in opposite ways:

\begin{itemize}[leftmargin=1.2em,itemsep=2pt,topsep=2pt]
\item \textbf{Symmetric noise attenuates.} With
$\rho_{F\to P}{=}\rho_{P\to F}{=}\rho$, the statistic is merely
\emph{compressed} toward $\tfrac12$ ($\kappa{=}1{-}2\rho$): its sign
is preserved, harmful skills still cross the threshold, and the only
cost is an inflated effective threshold $\tau/(1{-}2\rho)$ and a
$N_{\min}{\propto}(1{-}2\rho)^{-2}$ sample budget. Degradation is
graceful for every $\rho<\tfrac12$.
\item \textbf{False-pass bias displaces.} With $\rho_{F\to P}>0$
(denoted $q$ in the experiments) the
statistic shifts \emph{up}, most for the worst skills, so retirement
fires only if $\bar p(s)\le(\pi_\tau-\rho_{F\to P})/(1-\rho_{F\to P})$.
This right-hand side hits zero at $\rho_{F\to P}=\pi_\tau=(1-\tau)/2$:
beyond it, under this modeled channel, \emph{no skill is retired at any
sample size}. A cliff, not a slope.
\end{itemize}

\noindent This yields a bias-aware floor (Prop.~1$'$) and three design
rules: \emph{(i)} noise is
survivable but bias is not past the cliff; \emph{(ii)} the lever
against bias is the threshold $\tau$, not more data; \emph{(iii)}
$\rho_{F\to P}$ is measurable offline, so an operator can read off
which side of the cliff they are on, which motivates the audit
next. The full channel algebra, proof, and an adversarial-coupling
remark are deferred to Appendix~\ref{app:theory}.

\section{Gate: what can each signal see?}
\label{sec:gate}

Before trusting any reward channel we audit it against constructed
ground truth: inject one known defect into a clean section, ask each
grader whether it noticed.

\textbf{Defect classes.} Five \emph{QC-visible} classes mirror the
deterministic checks (inject an orphan citation; an unregistered
metric tag; a broken cross-reference; an unsourced number; delete the
TL;DR). Two \emph{QC-invisible} classes corrupt semantics while
preserving all annotation syntax: \emph{claim negation} (flip the
direction of a cited claim, ``growth'' $\to$ ``decline'') and
\emph{number swap} (perturb a digit while keeping the citation
marker on the line, so the sentence now misquotes its own source).

\begin{figure}[tbp]
\centering
\includegraphics[width=\textwidth]{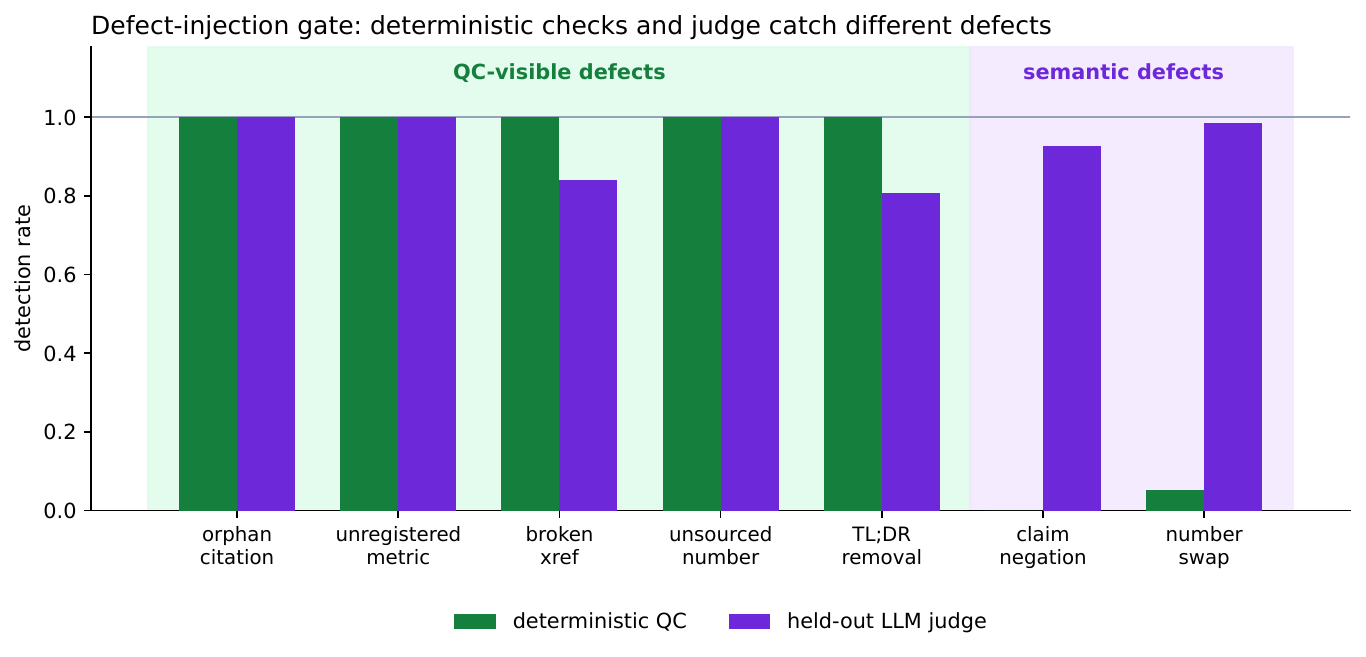}
\caption{Flag rate on injected defects (green = deterministic QC;
purple = held-out LLM judge). Bars are \emph{flag rates} (did the
grader raise any concern), not the scalar training score. QC catches
the five \emph{QC-visible} classes (recall $1.0$) but is blind to the
two semantic ones, which the judge flags. On structural defects the
judge often flags yet its scalar \emph{quality score} barely moves
($p\approx0.5$), so it does not reliably penalize them. The semantic
region, gradable only by a fallible judge, is the regime we study.}
\label{fig:gate}
\end{figure}

\textbf{Results} (155 sections, one injection per class per section).
The deterministic grader catches every QC-visible injection (recall
$1.0$, $n{=}155$ per class) and essentially none of the QC-invisible
ones (claim negation $0.0$; number swap $0.05$). A held-out judge (a
different model family, blind to condition, paired) flags both
QC-invisible defects at high rates (number swap $98.5\%$; claim
negation $92.7\%$; Fig.~\ref{fig:gate}), with significant
grounding-score drops (number swap $-1.15$, $p<10^{-4}$; claim negation
$-0.35$, $p{=}0.001$): precisely the defects the checks miss.
Conversely, on \emph{structural} defects the judge's quality score
barely moves (broken xref, TL;DR: $p\approx0.5$), which the checks
catch with certainty. The two signals are complementary: the checks do
not see meaning, the judge does not reliably penalize contract
violations (Fig.~\ref{fig:demo} in Appendix~\ref{app:demo} shows one
case verbatim). Their union
is the audit.

\textbf{Auditing the reward judge.} The same machinery measures the
error rates of the binary PASS/FAIL judge we later use as a
\emph{training reward} (Sec.~\ref{sec:sweep}), placing it on the
theory's axes: a strict, well-instructed judge has a tiny false-pass
rate ($\rho_{F\to P}\approx0.01$) but a large false-fail rate
($\rho_{P\to F}\approx0.95$). It thus sits not in the dangerous
false-pass corner but in the \emph{conservative} one, where
Prop.~1$'$ predicts safety bought at the price of \emph{starvation}
($\kappa\approx0.04$: almost no resolution to tell good skills from
bad). The realistic failure mode of a strict judge is signal
collapse, not reward hacking; the false-pass cliff is reached by
\emph{lenient} judges (or judges facing defects they cannot see).
This fixes the two roles corruption plays here. We \emph{inject}
false-pass bias on top of the deterministic reward to isolate its
causal channel, not because our particular judge is lenient: this one
is not. A judge's operating point is a measurable property rather than
a given, and lenient regions are easy to enter (a softer rubric, a
capable composer whose errors look fluent, or any defect the judge
cannot see). Audit details and a direction check ruling out
reward/quality conflict are in Appendix~\ref{app:gate}. The audit is
the practical payoff, a before-deployment go/no-go test that we
assemble into a deployment playbook once the empirical picture is
complete (Fig.~\ref{fig:playbook}, end of Sec.~\ref{sec:crossdomain}).

\section{The reward-reliability frontier}
\label{sec:sweep}

We run the loop for 12 rounds ($\tau{=}0.10$, $N_{\min}{=}24$,
$C{=}12$) under each reward channel: clean QC, symmetric noise, false-pass
bias $q$ (bracketing the predicted cliff at $0.45$), the audited LLM
judge, and a no-skill floor. Corruption hits \emph{training} rewards
only; evaluation always uses the true grader (Table~\ref{tab:sweep}).
We read the result at three levels: the \emph{mechanism} (does the
curator still retire?), the \emph{outcome} (does eval quality move?),
and the real LLM \emph{judge}'s place on this map.

\begin{table}[tbp]
\centering
{\small
\begin{tabular}{lccccccc}
\toprule
Condition & Tail eval & $\Delta$ clean & Synth & Dep. & True-ret. & Evict & Div. \\
\midrule
no-skill floor      & $0.628 \pm .059$ & $+0.060$ & $2$            & $0$ & $\mathbf{0}$           & $0$  & $0.00$ \\
clean QC            & $0.568 \pm .054$ & $0.000$  & $22.0 \pm 1.0$ & $9.7$ & $\mathbf{1.3 \pm 1.2}$ & $8.7$ & $0.00$ \\
\rowcolor{noiseband} noise $\rho{=}0.1$  & $0.601 \pm .049$ & $+0.033$ & $22.0 \pm 0.0$ & $10.0$ & $\mathbf{0.7 \pm 0.6}$ & $9.3$ & $0.10$ \\
\rowcolor{noiseband} noise $\rho{=}0.2$  & $0.586 \pm .054$ & $+0.018$ & $23.0 \pm 1.0$ & $11.0$ & $\mathbf{1.0 \pm 1.7}$ & $10.0$ & $0.22$ \\
\rowcolor{noiseband} noise $\rho{=}0.3$  & $0.595 \pm .010$ & $+0.027$ & $23.0 \pm 0.0$ & $11.0$ & $\mathbf{1.0 \pm 1.0}$ & $10.0$ & $0.31$ \\
\rowcolor{noiseband} noise $\rho{=}0.4$  & $0.628 \pm .063$ & $+0.060$ & $24.0 \pm 0.0$ & $12.0$ & $\mathbf{1.0 \pm 1.0}$ & $11.0$ & $0.38$ \\
\rowcolor{biasband} bias $q{=}0.2$      & $0.548 \pm .005$ & $-0.021$ & $19.7 \pm 1.2$ & $7.7$ & $\mathbf{0.0}$ & $7.7$ & $0.07$ \\
\rowcolor{biasband} bias $q{=}0.45$     & $\mathbf{0.503 \pm .036}$ & $\mathbf{-0.065}$ & $15.7 \pm 1.2$ & $3.7$ & $\mathbf{0.3 \pm 0.6}$ & $3.3$ & $0.18$ \\
\rowcolor{biasband} bias $q{=}0.7$      & $0.607 \pm .054$ & $+0.039$ & $15.0 \pm 1.0$ & $3.0$ & $\mathbf{0.0}$ & $3.0$ & $0.31$ \\
\rowcolor{judgeband} LLM judge          & $0.592 \pm .051$ & $+0.024$ & $19.0 \pm 1.7$ & $10.3$ & $\mathbf{10.3 \pm 3.2}$ & $0.0$ & $0.59$ \\
\bottomrule
\end{tabular}}
\caption{Reward-degradation sweep on \textsc{Report-main-71}
(mean\,$\pm$\,sd, 3 seeds; rows shaded blue = noise, red = bias,
purple = judge). Columns: \emph{Tail eval} = true-QC eval pass@1 over
the last 4 rounds; \emph{$\Delta$ clean} = vs the clean-reward loop;
\emph{Synth} = skills created; \emph{Dep.} = total deprecations,
$\approx$ \emph{True-ret.} (genuine contribution-retirement) $+$
\emph{Evict} (cap-evictions of healthy skills); \emph{Div.} = realized
corruption (fraction of training labels actually flipped, in
$[0,1]$). The bold \emph{True-ret.} column is the mechanism
signature: bias drives it to zero while noise and the real judge keep
it alive, even though the raw \emph{Dep.} count never reaches zero,
because eviction churn fills in.}
\label{tab:sweep}
\end{table}

\begin{figure}[tbp]
\centering
\includegraphics[width=\textwidth]{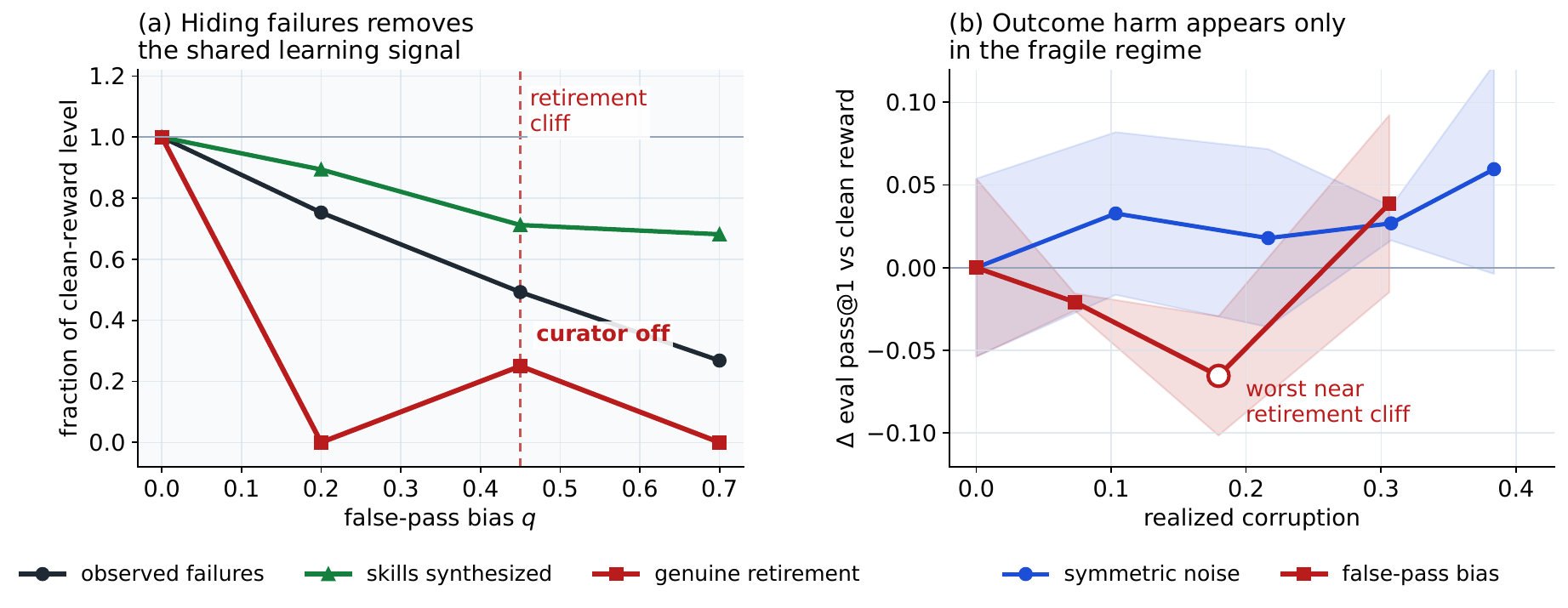}
\caption{\textsc{Report-main-71} (Claude Haiku 4.5, 3 seeds), as a
fraction of the clean-reward level. \textbf{(a) Mechanism.} Observed
failures are the loop's single input; rising false-pass bias $q$ shrinks
that pool and starves \emph{both} downstream stages, synthesis and
genuine retirement (${\approx}0$ at every bias rate here). \textbf{(b)
Outcome.} Eval damage vs the clean-reward loop, against realized
corruption (the fraction of training labels actually flipped):
false-pass bias is an inverted-U worst near the cliff
$q{=}(1-\tau)/2{=}0.45$, while symmetric noise stays at or above the
clean loop.}
\label{fig:main}
\end{figure}

\textbf{Mechanism level: genuine retirement, separated from churn.}
The mechanism column is \emph{True-ret.}, genuine contribution-based
retirement; we report it separately from cap-eviction (\emph{Evict})
because the raw deprecation count (\emph{Dep.}) mixes the two and only
the former reflects the curator's judgment (Sec.~\ref{sec:crossdomain}).
Reading \emph{Dep.} alone is misleading: it stays near $10$ under both
noise and the judge and never reaches zero under bias, so it hides a
curator that has stopped.
On \textsc{Report-main-71} the composer rarely produces a skill bad
enough to retire, so the clean baseline is already low ($1.3$), but
the asymmetry is exact (Fig.~\ref{fig:main}a): false-pass bias drives
true retirement to \emph{zero} at every rate ($0/0.3/0$ for
$q{=}0.2/0.45/0.7$), while symmetric noise holds it near the clean
level ($0.7$--$1.0$). The same corrupted signal also starves
failure-driven synthesis as bias grows (\emph{Synth} column,
$22\!\to\!15$). The
sharpest read is the \emph{judge} row: the realistic conservative judge
keeps true retirement fully alive ($10.3$, far above clean) precisely
because its false-pass rate is near zero, exactly the contrast
Prop.~1$'$ predicts. (Where the composer leaves more headroom the same
asymmetry shows with a larger baseline; Sec.~\ref{sec:replication}.)

\textbf{Outcome level: harm is worst at \emph{moderate} bias.}
Eval outcomes (vs the clean-reward loop) show the same asymmetry with
one wrinkle the simple cliff story misses (Fig.~\ref{fig:main}b). Noise
never hurts: it sits at or above the clean loop throughout ($+0.018$ to
$+0.060$ across rates; heavier noise churns the library more but the
curator keeps it healthy). Bias at
$q{=}0.2$/$0.45$ produces the worst outcomes in the sweep
($-0.021$/$-0.065$, and among the tightest across seeds): enough
failure signal survives to keep \emph{synthesizing} skills
($19.7$/$15.7$), but contribution-retirement can no longer weed them
(Fig.~\ref{fig:heatmap}, Appendix~\ref{app:heatmap}), so low-quality
skills accumulate to the cap and keep being routed to.
At extreme bias ($q{=}0.7$), however, the loop \emph{starves}:
with most failures reported as passes, synthesis itself (also
failure-driven) slows to $15.0$ skills and the near-inert library
drifts back to the clean level ($+0.039$). Harm is an inverted-U in
$\rho_{F\to P}$, peaking at the theory's
retirement-inoperative point $(1-\tau)/2 = 0.45$: the dangerous
judge is not the blindest one but the half-blind one, which feeds
the synthesizer while disarming the curator. The theory predicts
the retirement cliff; the inverted-U is what that cliff looks like
in a system where skill \emph{creation} shares the corrupted signal.

\textbf{The strict judge behaves as audited.} As its
$\rho_{F\to P}\approx0.01$ predicts, the real LLM judge, despite
the highest realized divergence ($0.59$, all phantom failures),
keeps retirement active and matches the clean loop ($+0.024$):
conservative error churns but does not disarm the curator. And the
absolute non-divergence the cap promises also held: no condition fell
more than a fixed margin below the no-skill floor (all means within
$0.125$ of it), including those where retirement was inoperative,
exactly as Prop.~1$'$ guarantees.

\textbf{Scope: harm, not lift.} On \textsc{Report-main-71} clean
evolution does not beat the no-skill floor ($-0.060$, under two of the
$28$ eval tasks): the composer is already strong, so the sweep measures
the \emph{differential damage of reward corruption}, not the loop's
upside. The actionable result is the harm ranking: against the clean
loop only moderate false-pass bias does real damage (worst at
$q{=}0.45$), while symmetric noise and the strict judge stay at or
above it. We replicate the mechanism in a harder, headroom-bearing
regime next.

\section{Replication in a harder regime, and a null lift result}
\label{sec:replication}

\textbf{The mechanism replicates with headroom.} To rule out a
near-ceiling artifact we re-run the whole sweep on
\textsc{Report-band-58}, a stricter reward on a bias-sensitive band
where the composer genuinely struggles (floor $0.388$;
Appendix~\ref{app:setup}, Fig.~\ref{fig:replication}). The signature is
sharper: the clean curator genuinely retires $\approx\!7$ skills/run,
noise keeps it there, and false-pass bias drives it to \emph{zero} by
$q{=}0.45$. The same noise-preserves / bias-kills pattern surviving a
change of composer difficulty, reward strictness, and floor level is
evidence it is a property of the governance mechanism, not one
operating point.

\textbf{No detectable end-to-end lift, and why that is the honest
finding.} Our claim is a replication of this \emph{mechanism}
(low-variance, seed-stable), not of an outcome lift. A single seed
suggested a $+0.12$ pass@1 lift, but it shrank
to $+0.014\pm0.054$ over three seeds: at $23$ binary-scored eval tasks
the variance is structural (binary scoring, unpaired means), so more
seeds cannot resolve it. We re-measured with a sharper instrument, a
\emph{paired} design scored on the \emph{continuous} violation count
(Table~\ref{tab:paired}, Appendix~\ref{app:paired}); the held-out lift
is still null (a mean reduction of just $0.26$ violations/section,
$p{=}0.63$; binary pass difference exactly $0$), as is the symmetric
harm test (clean vs bias-$0.45$ library, $p{=}0.61$). We claim the lift is
\emph{undetectable} at this
resolution, not provably zero. The one effect that survives is
mechanism-aligned: the library significantly reduces the single
violation class its skills police (unsourced numbers, full-band paired
$p{=}0.03$), confirming the skills do their narrow job.

\textbf{Why the micro-effect does not aggregate is itself the
finding.} A significant per-class improvement leaving headline quality
unmoved is the positive dual of this paper's thesis: our central
\emph{failure} (a disabled curator) and this \emph{success} (one
violation class driven down) are both invisible to aggregate eval,
because gains in one discipline are offset by others and the binary
pass collapses them all. That is why governance must be sized and
audited on the process signal that drives it, not the outcome metric:
the value is preventing degradation, as a non-divergence guarantee
promises, not manufacturing lift.

\section{Generality: universal mechanism failure, regime-dependent harm}
\label{sec:crossdomain}

Is the effect specific to long-form generation, or a property of the
curator's arithmetic that should appear wherever such governance runs?
We test generality along two axes: \emph{failure abundance} (the scarce
\textsc{Report-band-58} vs the abundant \textsc{Report-hard-133}, same
domain) and \emph{domain} (\textsc{MBPP+ hard100} code generation, a
\emph{perfect} unit-test verifier and a Claude Opus 4.7 composer, with
the same channels injected on the true pass/fail). This also forces a
sharper look at what ``retirement'' means.

\begin{figure}[tbp]
\centering
\includegraphics[width=\textwidth]{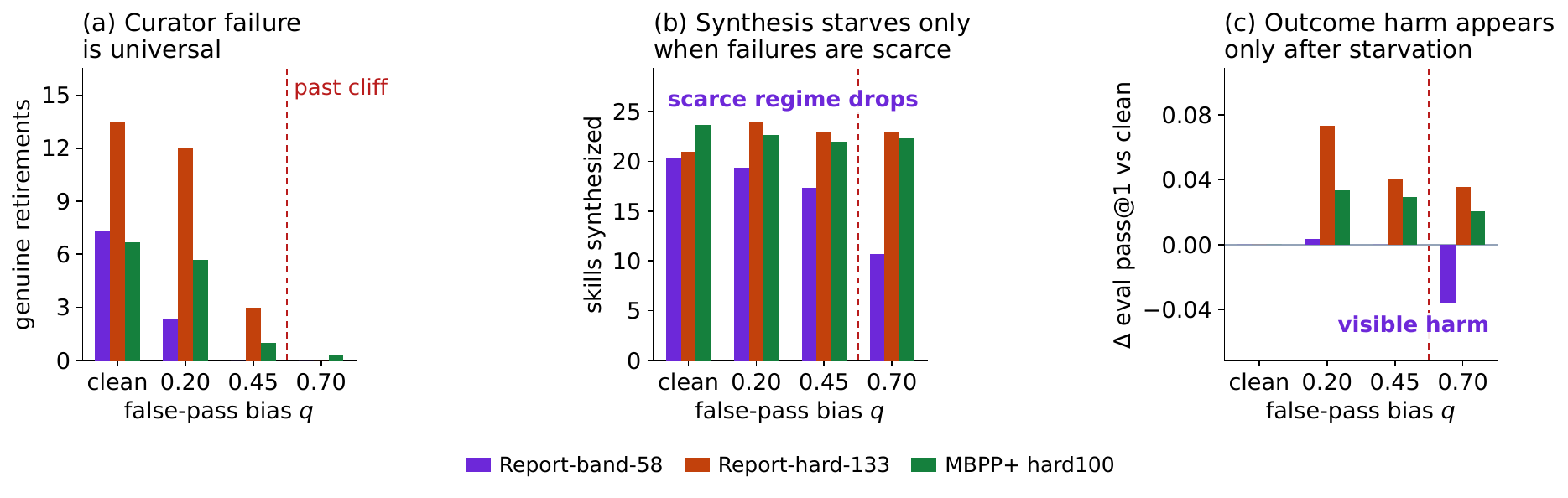}
\caption{The causal chain behind \emph{silent} curator failure, across
three subsets: \textsc{Report-band-58} (scarce failures),
\textsc{Report-hard-133} (abundant, same domain), and
\textsc{MBPP+ hard100} (abundant, different domain and model).
\textbf{(a)} genuine retirement falls to ${\approx}0$ past the cliff in
\emph{every} subset, the curator dies universally. \textbf{(b)}
synthesis survives where failures stay abundant and starves only in the
scarce subset. \textbf{(c)} so the eval outcome (vs clean) holds despite
the dead curator in the abundant subsets and degrades only where
synthesis starved: the curator's failure is \emph{silent} wherever
failures are plentiful.}
\label{fig:crossdomain}
\end{figure}

\textbf{Genuine retirement collapses everywhere.} Measured as
\emph{true} contribution-retirement (a skill whose observed
contribution reaches $-\tau$ at $\ge N_{\min}$ trials, separated from
the bounded bank's cap-evictions), false-pass bias drives retirement to
essentially zero at $q{=}0.7$ in \emph{every} subset
(Fig.~\ref{fig:crossdomain}a), including \textsc{MBPP+ hard100} ($0.3$)
and the abundant-failure \textsc{Report-hard-133} ($0.0$), not just the
scarce \textsc{Report-band-58}.
The realistic LLM judge, by contrast, keeps true retirement fully alive
(its near-zero false-pass rate, Sec.~\ref{sec:gate}), exactly as
Prop.~1$'$ predicts. So contribution-retirement is disabled by
false-pass bias as a \emph{universal} property of the curator's
arithmetic, independent of domain or failure abundance.

\textbf{Why the outcome diverges: synthesis, not retirement.} With the
curator dead everywhere, the downstream eval is set by whether the
\emph{other} failure-driven stage survives. Bias shrinks the failure
pool both stages feed on, but the \emph{absolute} failure volume
differs by subset: \textsc{Report-band-58}
falls from $20$ to $6$ failures/round and synthesis starves
($20\!\to\!11$, Fig.~\ref{fig:crossdomain}b), while the abundant
subsets still see $12$--$16$ and keep synthesizing at full rate
($\approx\!22$). The eval outcome tracks exactly this
(Fig.~\ref{fig:crossdomain}c): it holds at or above the clean loop in
both abundant subsets despite the dead curator and degrades
($-0.036$) only in the starved \textsc{Report-band-58}. This is
precisely why a disabled curator is \emph{silent} where failures are
plentiful: the outcome looks healthy while governance has quietly
stopped.

\begin{figure}[tbp]
\centering
\includegraphics[width=\textwidth]{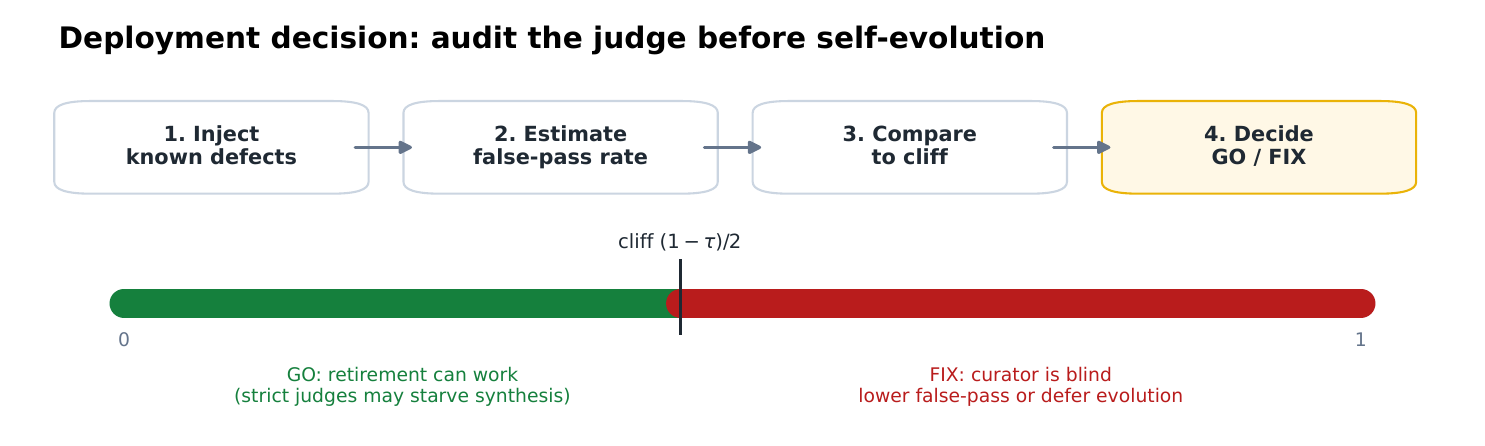}
\caption{Deployment playbook, assembling the paper's findings into a
before-deployment recipe. Audit the judge's false-pass rate
$\rho_{F\to P}$ offline (Sec.~\ref{sec:gate}); below the cliff
$(1-\tau)/2$ retirement works (a strict judge may starve synthesis),
above it the curator is blind, so lower the effective $\rho_{F\to P}$
or defer self-evolution.}
\label{fig:playbook}
\end{figure}

\textbf{Why this matters, and where.} The danger axis is the judge's
false-pass rate, not the domain. The domains where this governance was
validated (code, with unit tests) are safe because their ``judge'' is
a sound verifier with a near-zero false-pass rate, not because failures
are abundant. The deployments expanding fastest are exposed by
construction (deep research, multi-document analysis, open-ended
agentic writing): with no reference answer the reward \emph{must} be an
LLM judge, whose false-pass rate is real and unknown until measured.
Those settings also raise the cost of an undetected miss, since they
produce long, confident artifacts readers rarely re-verify. The regime
where our effect bites is thus both the
one verifier-free deployments are rushing into and the one where a
silently disabled curator does the most damage, which is why a one-time
judge audit is worth running first (the go/no-go playbook,
Fig.~\ref{fig:playbook}).

\section{Limitations}
\label{sec:limits}

Our scope is bounded. \emph{Coverage.} The study centers on one domain
and composer (MBPP+ adds a second of each as a control); the other
reference-free settings where the mechanism should next be tested (web
navigation, tool-use planning, multi-agent coordination) are untouched
here, and our seed-stable result is the synth/retire \emph{mechanism}
and its bias threshold, not an end-to-end lift (undetectable under
paired re-measurement; Appendix~\ref{app:paired}).
\emph{Exogenous corruption.} Imposing the channel on a deterministic
grader is what isolates its causal effect, so we can locate a real
judge relative to the cliff (Sec.~\ref{sec:gate}) but never exhibit one
that has drifted past it; a learned judge whose blindness the library
could exploit is out of scope (Appendix~\ref{app:theory}).
\emph{Model.} The channel is binary and the retirement rule attributes
each outcome to a single skill, so continuous rewards, multi-skill
composition, and learned retrievers change the \emph{estimator} rather
than the channel. \emph{Audit transfer.} Injection needs mutations that
provably change true quality; citation and numeric discipline make such
mutations cheap, and less localized failures do not.
Appendix~\ref{app:scope} treats both at length: what the channel model
covers, and what a task must supply for the audit to port to it.
Finally, the QC reward covers citation discipline, not insight, so a
section can pass QC yet be vacuous.

\section{Conclusion}

Failure-driven skill evolution survives a noisy judge, is disarmed by
a half-blind one, and is safe (though starvation-prone) under a strict
one, all distinguishable \emph{before} deployment by a cheap
defect-injection audit. The payoff is diagnostic, not numerical: no
aggregate pass rate separates those three regimes, so what the result
buys is the ability to see a governance failure, not a higher score.
Governance sized to the error structure of the signal that drives
it is the practical path for self-improving agents where no one can
write the unit test.

\input{main.bbl}
\appendix

\section{Evidence slice: schema and example}
\label{app:slice}

A trial gives the composer one \emph{evidence slice}: the section
brief plus the closed set of evidence \emph{cards}, metric tags, and
allowed cross-references it may use. The composer must cite only these
cards, annotate every numeral with a card or a registered metric tag,
hedge any card marked \texttt{status: weak}, and meet the structural
contract (e.g., a TL;DR with the required bullets). The example below
is synthetic (invented entity, fabricated figures); the real slices are
internal production data and follow the identical schema.

\noindent\begin{minipage}{\textwidth}
\begin{lstlisting}[basicstyle=\ttfamily\small,breaklines=true,
  columns=fullflexible,frame=single,framesep=4pt,xleftmargin=2pt]
{
  "section": "s3", "title": "Market position",
  "purpose": "size the addressable market and flag uncertainty",
  "tldr_min_bullets": 2,
  "cards": [
    { "id": "E001", "stance": "bull", "status": "ok",
      "claim": "ACME's FY25 revenue was $1.2B, up 30% YoY.",
      "quote": "...revenue reached $1.2B (+30% YoY)...",
      "source": "ACME FY25 annual report", "metric_refs": ["m_rev"] },
    { "id": "E002", "stance": "bear", "status": "weak",
      "claim": "A trade-press note estimates 2026 share near 18%.",
      "quote": "...we estimate ACME at ~18% share...",
      "source": "industry newsletter (secondary)" }
  ],
  "metrics": [ { "tag": "m_rev", "value": "1.2", "unit": "B USD" } ],
  "xref": [ { "id": "s5", "title": "Competitive landscape" } ]
}
\end{lstlisting}
\end{minipage}

\noindent A compliant section might open: ``\textbf{TL;DR} -- ACME
posted \texttt{\{m\_rev\}} revenue, +30\% \texttt{[E001]}; -- a
secondary estimate puts 2026 share near 18\% \texttt{[E002]}, though
this figure is unverified (see \S s5).'' Here \texttt{[E001]} and
\texttt{[E002]} are valid citations, \texttt{\{m\_rev\}} is a
registered metric tag, the weak card is hedged, and the TL;DR meets
its bullet floor: deterministic QC passes. This exposes the two defect
regimes concretely. A \emph{QC-visible} defect breaks the
syntax, e.g., citing \texttt{[E003]} (not in the slice, an orphan
citation) or writing a bare ``30\%'' with no tag: the checks catch it
with certainty. A \emph{QC-invisible} defect preserves every marker
but corrupts meaning, e.g., flipping ``+30\%'' to ``$-$30\%'' while
keeping \texttt{[E001]}: the citation still resolves, so QC passes,
yet the sentence now misquotes its own source. Only a judge that reads
the quote can catch the second kind, which is the gap
Sec.~\ref{sec:gate} measures.

\section{Experimental details}
\label{app:setup}

Table~\ref{tab:experiments} collects every experiment in the paper,
its subset, composer, reward, size, and where it is used; the rest of
this appendix gives only what does not fit the table.

\begin{table}[t]
\centering
{\small
\setlength{\tabcolsep}{4pt}
\begin{tabular}{lllccll}
\toprule
Name & Domain, composer & Reward & tr/ev & fail & floor & Used in \\
\midrule
\textsc{Report-main-71} & report, Haiku 4.5 & QC (5 chk) & 43/28 & 0.41 & 0.628
  & Tab.~\ref{tab:sweep}, Fig.~\ref{fig:main} \\
\textsc{Report-band-58} & report, Haiku 4.5 & QC$+$tier-2 & 35/23 & 0.62 & 0.388
  & Fig.~\ref{fig:replication}, Tab.~\ref{tab:paired} \\
\textsc{Report-hard-133} & report, Haiku 4.5 & QC$+$tier-2 & 80/53 & 0.74 & 0.198
  & Fig.~\ref{fig:crossdomain} \\
\textsc{MBPP+ hard100} & code, Opus 4.7 & unit tests & 60/40 & 0.67 & 0.273
  & Fig.~\ref{fig:crossdomain} \\
\textsc{Report-weak-71} & report, 3.5 Haiku & QC (5 chk) & 43/28 & n/a & 0.17
  & \S\ref{app:setup} \\
\bottomrule
\end{tabular}}
\caption{All experiments at a glance. Common to every run: 3 seeds
(42/43/44) $\times$ 12 rounds, $\tau{=}0.10$, $N_{\min}{=}24$, cap
$C{=}12$; the reward channels (clean / symmetric noise
$\rho\in\{.1,.2,.3,.4\}$ / false-pass bias $q\in\{.2,.45,.7\}$ /
audited LLM judge) are injected on the \emph{training} reward only,
while evaluation always uses the true grader. \emph{tr/ev} = train /
eval sections; \emph{fail} = clean true-failure rate on the train
pool; \emph{floor} = no-skill tail eval pass@1.}
\label{tab:experiments}
\end{table}

\begin{figure}[tbp]
\centering
\includegraphics[width=\textwidth]{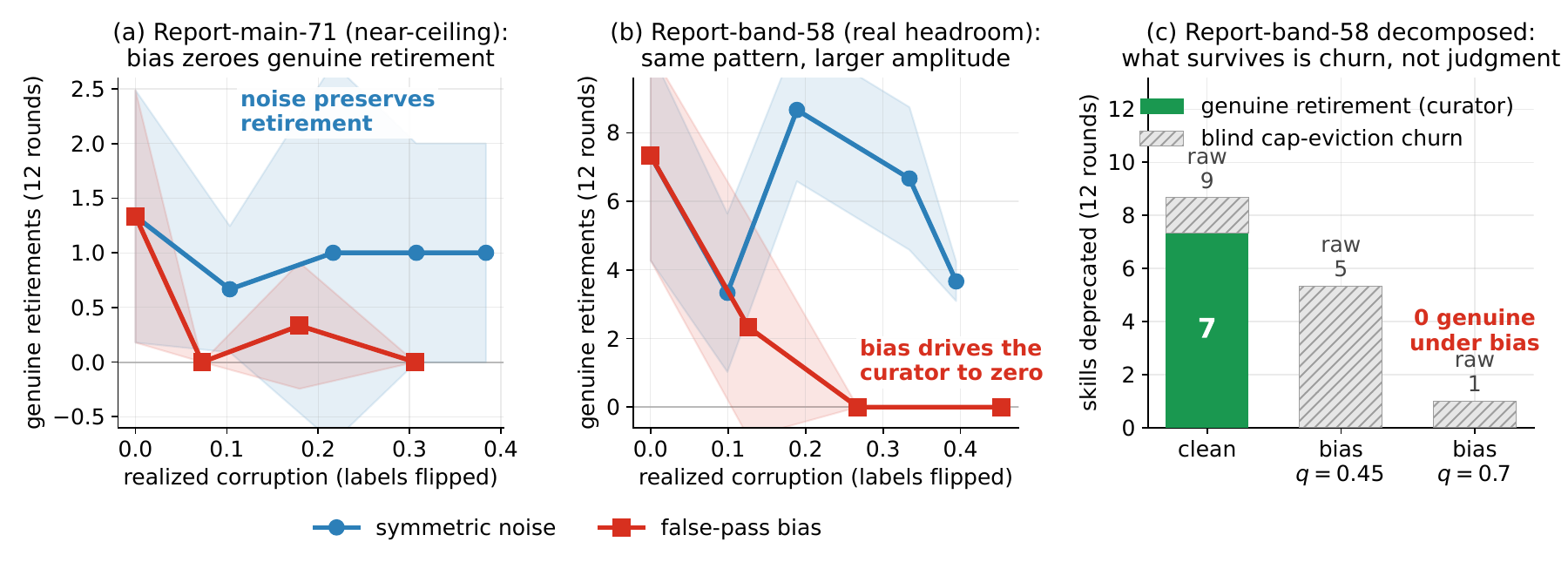}
\caption{The retirement signature replicates across two subsets (3
seeds; blue = noise, red = false-pass bias, vs realized corruption, the
fraction of training labels actually flipped).
\textbf{(a)} \textsc{Report-main-71} (near-ceiling): noise preserves
genuine retirement, bias drives it to zero. \textbf{(b)}
\textsc{Report-band-58} (real headroom): same pattern, larger amplitude
($\approx\!7\!\to\!0$ by $q{=}0.45$). \textbf{(c)} decomposing the
\textsc{Report-band-58} deprecation count shows genuine retirement
(green) collapsing $7\!\to\!0$ while blind cap-eviction churn (gray)
props up the raw total.}
\label{fig:replication}
\end{figure}

\textbf{Models and inference.} Composers are sampled at temperature
$0.7$ with a $2200$-token cap. The in-loop judge reward channel uses
the same model as the composer, scored deterministically at
temperature $0$ (binary \textsc{pass}/\textsc{fail}, $200$-token cap).
The gate audit (Fig.~\ref{fig:gate}) uses a held-out judge of a
\emph{different} family (Claude Sonnet 4.6, temperature $0$) so its
errors are not shared with the composer. Noise and false-pass bias are
injected synthetically on the training reward label only; the
corruption never touches inference.

\textbf{Subset construction.} The report subsets keep only sections
with genuine headroom. \textsc{Report-main-71} keeps the 71 of 155 that
fail at least once. \textsc{Report-band-58} keeps the 58 the composer
fails on $1$ or $2$ of $3$ probes, excluding always-pass sections (no
headroom) and always-fail ones (unfixable), and adds tier-2 content
disciplines (hedge weak-status cards, cover both stances, meet a
coverage floor) to the reward for real headroom.
\textsc{Report-hard-133} adds the always-fail sections back, raising the
true-failure rate and synthesis pressure without changing the reward: a
same-domain, failure-abundant control where (as in MBPP) the headline
deprecation count stays high under bias but is cap-eviction churn,
while genuine contribution-retirement still collapses to zero
(Fig.~\ref{fig:crossdomain}). \textsc{MBPP+ hard100} uses Ratchet's
published hard-100 split unchanged, with the channels injected on the
true unit-test pass/fail (baseline pass@1 $0.273$, vs $0.258$ originally
reported).

\textbf{\textsc{Report-weak-71} (capability control).} It swaps in a
weaker composer (Claude 3.5 Haiku) on the main subset, leaving ample
headroom (floor $\approx0.17$) yet clean evolution does not beat it
(tail $0.171$, equal to floor): skills amplify a capable composer
rather than teach a weak one, so \textsc{Report-band-58} raises
difficulty through the reward instead of weakening the model.

\input{theory}

\section{A QC-invisible defect, verbatim}
\label{app:demo}

\begin{figure}[t]
\centering
\includegraphics[width=\textwidth]{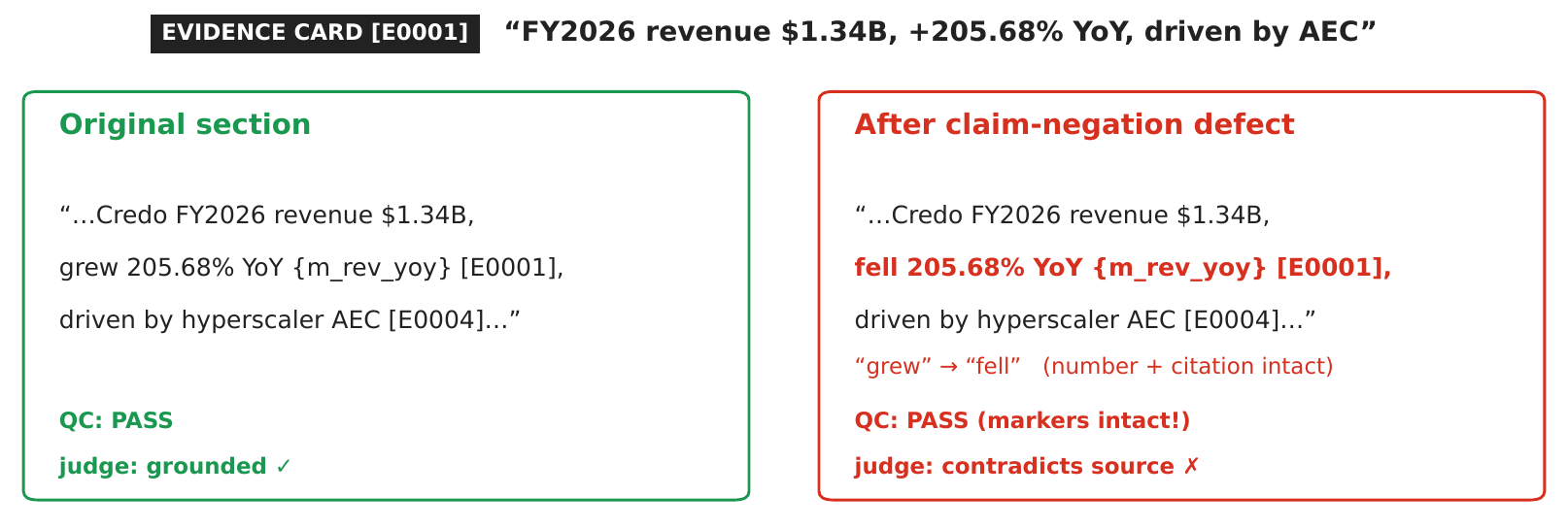}
\caption{A real \emph{claim-negation} defect (demo-credo): flipping
the direction word (``grew''\,$\to$\,``fell'') leaves the number
($205.68\%$), metric tag \texttt{\{m\_rev\_yoy\}}, and citation
\texttt{[E0001]} byte-identical, so deterministic QC (and even a value
check) sees nothing wrong, yet the sentence now contradicts its own
cited source. The QC-invisible / judge-visible gap in one example.}
\label{fig:demo}
\end{figure}

Figure~\ref{fig:demo} shows one injected defect at the granularity the
injector operates on, an English gloss of a section from one of the
deep-research productions of Sec.~\ref{sec:setup}. It is the concrete
case behind the two error classes of
Sec.~\ref{sec:gate}: what the deterministic checks read (markers,
metric tags, numbers) is untouched, so the mutation is invisible to the
reward we corrupt, and only a semantic grader has anything to detect.

\section{Gate audit details}
\label{app:gate}

\textbf{Direction check.} On clean sections, judge scores do not
anti-correlate with deterministic warning counts (Spearman
$r_s{=}0.19$, $95\%$ CI $[0.03,0.34]$, $n{=}143$): a weak positive,
consistent with number-dense sections being both warning-prone and
information-rich. This rules out the failure mode in which optimizing
the QC reward would actively fight judge-perceived quality.

\textbf{Reward-judge audit.} We grade fresh compositions from the
sweep's composer with defects injected only into truly-passing
sections, giving $\rho_{F\to P}\approx0.01$ on QC-visible defects
($n{=}210$) and $0.014$ on QC-invisible ones ($n{=}70$), against
$\rho_{P\to F}\approx0.95$ on truly-passing sections ($n{=}42$):
i.e., $\kappa\approx0.04$, the conservative-corner regime discussed in
Sec.~\ref{sec:gate}.

\section{Curation activity over time}
\label{app:heatmap}

\begin{figure}[t]
\centering
\includegraphics[width=\textwidth]{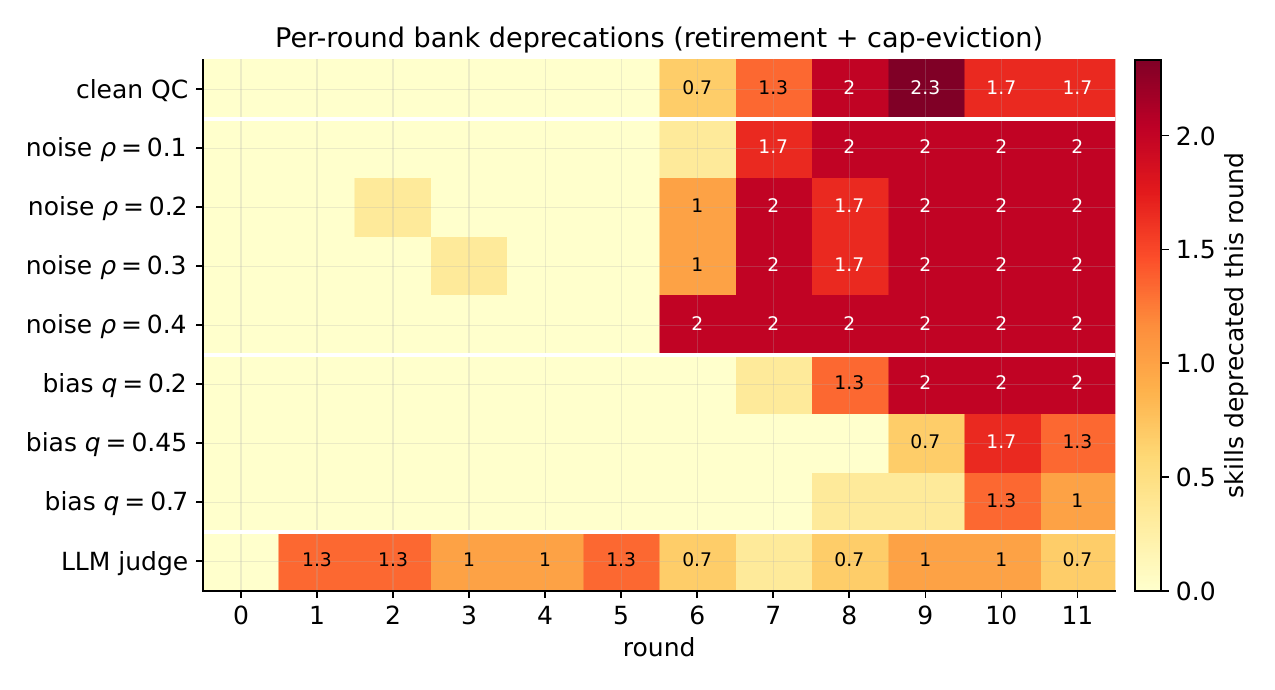}
\caption{Per-round bank deprecations (mean of 3 seeds), \textsc{Report-main-71}:
each row a condition, each column a round, color = skills deprecated
that round (contribution-retirement plus cap-eviction). After the
$N_{\min}$ warm-up the bank stays active under noise but goes
increasingly quiet under false-pass bias (pale, delayed): the
curation loop winding down in time.}
\label{fig:heatmap}
\end{figure}

This temporal view complements the aggregate counts of
Sec.~\ref{sec:sweep} (Table~\ref{tab:sweep}): the loop quiets round by
round under bias, not all at once.

\section{Paired, continuous lift measurement}
\label{app:paired}

This appendix details the re-measurement summarized in
Sec.~\ref{sec:replication}. The end-to-end lift reported by a single
seed ($+0.12$ pass@1) failed to survive three seeds
($+0.014\pm0.054$), and we trace this to the \emph{measurement}, not
the sample size: with $23$ binary-scored eval tasks the quantum is
$1/23\approx0.043$ and the dominant variance comes from comparing two
unpaired noisy means across sections of very different difficulty. No
number of seeds removes that structural variance.

\textbf{Protocol.} We freeze the final evolved library from each
\textsc{Report-band-58} run and, for every section, compose it twice with the
same Claude Haiku 4.5 composer at the eval temperature: once with the section
routed to the frozen library, once with routing forced to the no-skill
path (the floor arm). Each cell is averaged over three composer
repeats. Scoring uses the same deterministic tier-2 grader, but on its
\emph{continuous} violation count rather than the binary
\textsc{pass} (the judge never enters evaluation). Pairing cancels
per-section difficulty, the largest noise source; the continuous score
raises resolution by an order of magnitude. We report Wilcoxon
signed-rank over the per-section paired differences, on the $23$
held-out eval sections (primary) and on all $58$ \textsc{Report-band-58}
sections (higher-power secondary, train sections in-sample).

\begin{table}[t]
\centering
{\small
\begin{tabular}{lcc}
\toprule
Measurement & held-out eval ($n{=}23$) & full band ($n{=}58$) \\
\midrule
Lift, continuous violations ($\Delta$) & $+0.26$ ($p{=}0.63$) & $+0.23$ ($p{=}0.15$) \\
Lift, binary pass@1 & $+0.00$ ($p{=}0.88$) & $+0.05$ ($p{=}0.15$) \\
Harm, continuous (clean vs bias) & $+0.23$ ($p{=}0.61$) & $+0.32$ ($p{=}0.22$) \\
\midrule
\emph{Targeted class} (unsourced number) & $+0.30$ ($p{=}0.08$) & $+0.22$ ($p{=}0.03$) \\
\bottomrule
\end{tabular}}
\caption{Paired lift on \textsc{Report-band-58}. $\Delta$ = per-section
reduction in violations from the evolved library vs the no-skill floor
(positive $=$ helps); pass@1 = binary-outcome difference. The
end-to-end lift is undetectable; the only significant effect is the
reduction of the \emph{specific} violation class the skills target.}
\label{tab:paired}
\end{table}

\textbf{Reading the table.} The harm direction (clean library vs the
bias-$0.45$ library) is null at the same resolution as the lift
($p{=}0.61$): at $23$ sections the aggregate eval is insensitive to
both, which is why this paper's evidence rests on the mechanism signal
(synth/retire, low-variance; Sec.~\ref{sec:sweep}) and on the
class-level signal instead. The one surviving effect is
mechanism-aligned, since the synthesized skills police citation and
sourcing discipline: the library reduces precisely the unsourced-number
violation class on the full band ($p{=}0.03$), while gains there are
offset by other disciplines so the aggregate does not move. Note also
what the design does and does not credit: it measures the value of the
\emph{final} library, not the evolution trajectory, which the
synth/retire results (Sec.~\ref{sec:sweep}) address directly.

\section{Scope of the channel model, and porting the audit}
\label{app:scope}

This appendix expands the \emph{Model} and \emph{Audit transfer} limits
of Sec.~\ref{sec:limits}. Nothing here is measured; it is a statement
of what the analysis covers and what a new domain has to supply.

\textbf{What the cliff depends on.} The displacement result needs only
three ingredients: a retirement rule that thresholds an observed
quality statistic, a fixed threshold, and a judge whose error has a
sign. Symmetric error rescales the statistic around its midpoint and
preserves order, so the rule still fires at an inflated margin. Signed
error moves the statistic bodily, and once the displacement exceeds the
distance from the worst attainable value to the threshold, the rule
cannot fire at any sample size. The particular constant
$\pi_\tau = (1-\tau)/2$ is Ratchet's: it is where a $\pm1$ contribution
scale puts the retirement cut. A different governance rule inherits the
same dichotomy with its own constant.

\textbf{Continuous rewards.} With a bounded continuous score in place
of a binary outcome, the retirement statistic is a mean score rather
than a pass rate, and the channel becomes an error distribution on that
score. A zero-mean error leaves the statistic's location alone and only
adds variance, which $N_{\min}$ absorbs; an error with non-zero mean in
the lenient direction shifts that location exactly as $\rho_{F\to P}$
does here.
So we expect the attenuate-versus-displace contrast to be about the
sign structure of judge error rather than about the reward's
cardinality. We state this as the reading of the mechanism, not as a
proved extension: the binary case is what Prop.~1$'$ covers and what
the sweep tests.

\textbf{Multi-skill composition and learned retrievers.} Ratchet routes
at most one skill per task, which is what makes a trial's outcome
attributable to a single skill. Compose several skills per trial, or
let a learned retriever decide which, and per-skill contribution stops
being identified from the outcome alone: the quantity the curator
thresholds is now the output of a credit-assignment procedure, and
judge bias compounds with attribution bias. Our analysis deliberately
holds attribution fixed and corrupts only the outcome, isolating the
cheaper of the two failure modes. A curator fed a biased judge
\emph{and} a confounded attribution is strictly worse off than the one
studied here, but the size of the effect is not something this paper
measures.

\textbf{What a task must supply for the audit.} Defect injection needs
(i) a mutation operator that changes true quality \emph{by
construction}, so ground truth is known without consulting any grader,
(ii) mutation classes on both sides of the checkable boundary, so the
audit separates what deterministic checks catch from what only a judge
can, and (iii) enough trials per class to estimate a rate. Report
composition makes (i) cheap because its failures localize: flipping a
direction word or swapping a citation identifier provably changes
faithfulness while leaving every syntactic feature intact
(Appendix~\ref{app:demo}).

Where failures do not localize (a plan that is subtly unachievable, a
navigation trajectory that ends in the wrong state), the practical
substitute is to mutate the \emph{task} rather than the output:
perturb a precondition so that a previously correct artifact becomes
wrong by construction, and the same known-ground-truth property holds.
Building that operator is the domain expertise the audit needs. It is a
one-time cost per domain, paid once against every subsequent
deployment, which is the reason we present the audit as a
before-deployment gate rather than as a monitoring metric.

\typeout{get arXiv to do 4 passes: Label(s) may have changed. Rerun}

\end{document}

%% file: theory.tex
\section{Full theory: bias-aware non-divergence}
\label{app:theory}

This appendix gives the full derivation summarized in
Sec.~\ref{sec:theory}: the corruption channel, the two
corruption regimes, and the bias-aware floor (Proposition~1$'$) with
proof. Ratchet's original guarantee (Prop.~1) assumes the per-skill
contribution estimator is unbiased; with a deterministic grader this
is innocuous, but in reference-free domains the grader is an LLM judge
whose errors are \emph{structured}, not mere variance. We trace that
structure through the retirement rule.

\subsection{Corruption channel}

Let $y \in \{0,1\}$ be the true outcome of one trial (as scored by a
perfect grader) and $\tilde{y}$ the observed outcome. Model the judge
as a binary channel,
\[
\Pr[\tilde{y}{=}1 \mid y{=}0] = \rho_{F\to P}, \qquad
\Pr[\tilde{y}{=}0 \mid y{=}1] = \rho_{P\to F},
\qquad \rho_{F\to P} + \rho_{P\to F} < 1,
\]
encoding $y{=}1$ as a true \textsc{pass} and $y{=}0$ as a true
\textsc{fail}. Subscripts read as \emph{true$\to$observed}:
$\rho_{F\to P}$ is the rate at which a true failure is observed as a
pass (a \emph{hidden} failure), and $\rho_{P\to F}$ the rate at which
a true pass is observed as a fail (a \emph{phantom} failure). We avoid
the ``false positive/negative'' labels deliberately: they invert
depending on whether one takes a pass or a failure as the positive
class, and only the hidden-failure direction $\rho_{F\to P}$ disarms
retirement.
Ratchet retires a skill $s$ when its empirical contribution
$\hat{c}(s) = (\,\#\mathrm{succ} - \#\mathrm{fail}\,)/n(s)
= 2\hat{p}(s) - 1$ falls to $-\tau$ after $n(s) \ge N_{\min}$
trials; equivalently, when the observed pass rate
$\hat{p}(s) \le \pi_\tau := \tfrac{1-\tau}{2}$. Under the channel,
the observed pass rate concentrates on
\[
p_{\mathrm{obs}}(s)
 = \bigl(1 - \rho_{F\to P} - \rho_{P\to F}\bigr)\,
   \bar{p}(s) + \rho_{F\to P}
 \;=\; \kappa\,\bar{p}(s) + \rho_{F\to P},
\qquad \kappa := 1 - \rho_{F\to P} - \rho_{P\to F},
\]
where $\bar{p}(s)$ is the skill's true pass rate on tasks routed to
it. Two consequences, one per corruption type:

\textbf{(i) Symmetric noise} ($\rho_{F\to P} = \rho_{P\to F}
= \rho < \tfrac12$): then $p_{\mathrm{obs}} - \tfrac12 =
(1-2\rho)(\bar{p} - \tfrac12)$, so the statistic is \emph{compressed}
toward $\tfrac12$ but its ordering is preserved. Retirement of a
truly harmful skill still fires, at an inflated margin: the true pass
rate must satisfy
$\bar{p}(s) \le \tfrac12 - \tfrac{\tau/2}{1-2\rho}$,
i.e., the effective retirement threshold is
$\tau_{\mathrm{eff}} = \tau/(1-2\rho)$, and Hoeffding's radius must
now resolve means separated by a factor $(1-2\rho)$, so the required
$N_{\min}$ grows as $(1-2\rho)^{-2}$. Degradation is \emph{graceful}:
finite for every $\rho < \tfrac12$.

\textbf{(ii) False-pass bias} ($\rho_{F\to P} > 0$,
$\rho_{P\to F} = 0$): then
$p_{\mathrm{obs}} = \bar{p} + \rho_{F\to P}(1 - \bar{p})$,
an additive \emph{displacement}, largest exactly for the worst
skills (small $\bar{p}$). Retirement fires only if
$\bar{p}(s) \le \frac{\pi_\tau - \rho_{F\to P}}
{1 - \rho_{F\to P}}$. The right-hand side hits $0$ at
$\rho_{F\to P} = \pi_\tau = \tfrac{1-\tau}{2}$: beyond this
point \emph{no skill is ever retired, at any sample size}, since more
trials concentrate the estimator more tightly around a displaced
mean. The mechanism predicts a \emph{cliff}, not a slope.

\subsection{\texorpdfstring{Proposition 1$'$}{Proposition 1'}}

\textbf{Proposition 1$'$ (Non-divergence under corrupted reward).}
Assume the Router conditions of Prop.~1 and the channel above with
known bounds $\rho_{F\to P} \le \bar\rho_{F\to P}$,
$\rho_{P\to F} \le \bar\rho_{P\to F}$,
$\kappa \ge \underline\kappa > 0$. Choose $N_{\min}$ so that the
observed pass rate of every ACTIVE skill is within $\epsilon$ of its
mean w.p.\ $\ge 1-\delta$ (Hoeffding). Then expected eval pass@1
under Ratchet is lower-bounded by
\[
\mathbb{E}[p_0]
\;-\;
\frac{\tau/2 \;+\; \epsilon \;+\; \bar\rho_{F\to P}}
     {\underline\kappa}
\;-\; \tfrac12\bigl(1 - \underline\kappa\bigr)
\;-\; C_0\,\delta
\]
(up to the affine map between pass-rate and contribution scales, with
$C_0$ an absolute constant).

\emph{Proof sketch.} On the high-probability event every surviving
skill has observed pass rate $\ge \pi_\tau - \epsilon$. Inverting
the channel, its true pass rate satisfies $\bar{p}(s) \ge
(\pi_\tau - \epsilon - \bar\rho_{F\to P})/\underline\kappa$.
Comparing against the NONE route's $\bar{p}_0$ and taking
expectations over the routing distribution reproduces the Prop.~1
argument with the inflated margin. \qed

\textbf{Reading the bound.} Three design rules fall out.
(1)~\emph{Noise attenuates, bias displaces}: symmetric noise enters
only through $\underline\kappa = 1-2\rho$ and the floor degrades
smoothly for all $\rho<\tfrac12$; false-pass bias enters additively
through $\bar\rho_{F\to P}$ and renders retirement inoperative
at $\bar\rho_{F\to P} \ge \pi_\tau$, where the bound goes
vacuous discontinuously.
(2)~\emph{Compensate with $\tau$, not with $N_{\min}$}: sample size
shrinks $\epsilon$ but never $\bar\rho_{F\to P}$; the only
lever against bias is widening the retirement threshold
($\pi_\tau > \bar\rho_{F\to P}$, i.e.,
$\tau < 1 - 2\bar\rho_{F\to P}$), which is possible only while
the judge's false-pass rate is below $\pi_\tau$.
(3)~\emph{Measure the judge, read off the floor}:
$\bar\rho_{F\to P}, \bar\rho_{P\to F}$ are estimable
offline by defect injection against constructed ground truth
(Sec.~\ref{sec:gate}), making the bound operational: audit the
judge once, then know which side of the cliff your evolution loop
sits on.

\textbf{Falsifiable prediction.} Library evolution driven by a
reward channel with symmetric noise $\rho$ should degrade gracefully
in $\rho$ and remain non-divergent up to high noise; evolution driven
by false-pass bias $\rho_{F\to P}$ should hold and then fail
abruptly near $\rho_{F\to P} \approx \pi_\tau$. The experiments
of Sec.~\ref{sec:sweep} test exactly this contrast.

\textbf{Remark (adversarial coupling).} If skill content can
\emph{cause} judge blindness (the library learns phrasing that
fabricates confidently), then $\rho_{F\to P}$ becomes
skill-dependent and grows along the evolution trajectory; no fixed
audit bounds it. Our sweep deliberately breaks this coupling
(corruption is injected exogenously on top of a deterministic
grader), isolating the channel's effect; the fully-learned-judge
condition, where the coupling is live, is out of scope here
(Sec.~\ref{sec:limits}).